\documentclass[letterpaper]{article}
\usepackage[preprint]{aaai2027}
\usepackage[hyphens]{url}
\usepackage{graphicx}
\usepackage{natbib}
\usepackage{caption}
\usepackage{booktabs}
\usepackage{multirow}
\usepackage{amssymb}
\usepackage{amsmath}
\usepackage[table]{xcolor}
\definecolor{n8col}{gray}{0.90}

\newcommand{\cmark}{\checkmark}

\title{Video Understanding Reward Modeling: A Robust Benchmark and Performant Reward Models}

\author{
    Yuancheng Wei\textsuperscript{\rm 1}\equalcontrib,
    Linli Yao\textsuperscript{\rm 2}\equalcontrib,
    Lei Li\textsuperscript{\rm 3}\equalcontrib,
    Haojie Zhang\textsuperscript{\rm 1},\\
    Hao Zhou\textsuperscript{\rm 4}\thanks{Project leader.},
    Fandong Meng\textsuperscript{\rm 4},
    Xu Sun\textsuperscript{\rm 2}\corresponding
}
\affiliations{
    \textsuperscript{\rm 1}South China University of Technology\quad
    \textsuperscript{\rm 2}Peking University\\
    \textsuperscript{\rm 3}The University of Hong Kong\quad
    \textsuperscript{\rm 4}Tencent\\
    wyc528813339@gmail.com\\
    Code: https://github.com/wyclike/VURM
}

\begin{document}

\maketitle

\begin{abstract}
Multimodal reward models have advanced in text and image domains, yet progress in video understanding reward modeling remains limited by the lack of robust evaluation benchmarks and high-quality preference data. To address this, we propose a unified framework spanning benchmark design, data construction, and reward model training. We introduce \textbf{Video Understanding Reward Bench} (VURB), a benchmark featuring 2,100 preference pairs with long chain-of-thought reasoning traces (averaging 1,143 tokens) and majority voting evaluation across general, long, and reasoning-oriented video tasks. We further construct \textbf{Video Understanding Preference Dataset} (VUP-35K) via a fully automated pipeline, providing large-scale high-quality supervision for video reward training. Building on the data, we train \textbf{VideoDRM} and \textbf{VideoGRM}, a discriminative and a generative reward model, both achieving state-of-the-art performance on VURB and VideoRewardBench. Further analysis confirms that VUP-35K enhances both reward performance and model reasoning capability, while both models yield significant gains under best-of-$N$ test-time scaling.

\end{abstract}

\section{Introduction}
\label{sec:intro}

\begin{table*}[t]
\centering
\footnotesize
\setlength{\tabcolsep}{3pt}
\renewcommand{\arraystretch}{1.15}
\begin{tabular}{lcccccc}
\toprule
\textbf{Dataset} & \begin{tabular}[c]{@{}c@{}}\textbf{Video Pref.}\\\textbf{Pairs}\end{tabular} & \begin{tabular}[c]{@{}c@{}}\textbf{Dim.}\\\textbf{Split}\end{tabular} & \begin{tabular}[c]{@{}c@{}}\textbf{All MRM}\\\textbf{Types}\end{tabular} & \begin{tabular}[c]{@{}c@{}}\textbf{Resp. With CoT}\\\textbf{Ratio}\end{tabular} & \begin{tabular}[c]{@{}c@{}}\textbf{Avg Resp.}\\\textbf{Len.}\end{tabular} & \begin{tabular}[c]{@{}c@{}}\textbf{Evaluation}\\\textbf{Protocol}\end{tabular} \\
\midrule
MM-RLHF-RewardBench~\cite{mmrlhf} & 25 & $\times$ & $\times$ & 0\% & 52 & vanilla \\
JudgeAnything~\cite{judgeanything} & 100 & $\times$ & $\times$ & - & 73 & Majority Voting \\
OmniRewardBench~\cite{omnireward} & 443 & $\times$ & \cmark & - & 133 & vanilla \\
VideoRewardBench~\cite{videorewardbench} & 1559 & \cmark & \cmark & 35.9\% & 104 & vanilla \\
\midrule
\textbf{VURB(Ours)} & \textbf{2100} & \textbf{\cmark} & \textbf{\cmark} & \textbf{100\%} & \textbf{1143} & \textbf{Majority Voting} \\
\bottomrule
\end{tabular}
\caption{\textbf{Comparison with previous benchmarks on video understanding data.} ``Dim. Split'' denotes dimensional splits, while ``All MRM Types'' indicates whether the benchmark evaluates various types of multimodal reward models. VURB provides dimension-level splits, evaluates all major MRM types, includes long chain-of-thought reasoning in every preference response, and employs majority voting for robust assessment.}
\label{tab:comparison}
\end{table*}

Multimodal reward models (MRMs) are central to modern training pipelines, supporting data filtering~\cite{alpagasus}, reinforcement learning~\cite{rm}, and test-time scaling~\cite{reinforced}. Despite rapid progress in text and image reward modeling~\cite{vlrewardbench}, video understanding reward modeling remains constrained by two bottlenecks that surface on both the evaluation and data sides: the absence of a robust benchmark and the scarcity of high-quality preference data.

On the evaluation side, as shown in Table~\ref{tab:comparison}, existing benchmarks suffer from limited scale, insufficient dimensional coverage, and unreliable evaluation protocols. Specifically, general multimodal reward benchmarks such as MM-RLHF-RewardBench~\cite{mmrlhf}, JudgeAnything~\cite{judgeanything}, and OmniReward~\cite{omnireward} contain few video understanding preferences and cover only a narrow range of video understanding abilities. VideoRewardBench~\cite{videorewardbench} expands the scale to 1,559 samples with task-specific dimensions, but its responses average merely 104 tokens and largely omit the chain-of-thought (CoT) paradigm~\cite{cot} prevalent in modern video understanding, providing inadequate logical context for effective preference judgment. Furthermore, its single-pass evaluation protocol is vulnerable to position bias~\cite{positionbias}. On the data side, prior preference learning has focused mainly on video \emph{generation}~\cite{videodpo}, while large-scale, high-quality preference data for video \emph{understanding} remains scarce because of high annotation costs. 
Together, the lack of a robust benchmark and large-scale, high-quality preference data creates a bottleneck for video reward modeling.

To address these limitations, we propose a unified framework covering benchmark design, preference data construction, and reward model training. Specifically, we introduce \textbf{Video Understanding Reward Bench} (VURB), a benchmark for video understanding preference judgment, spanning general, long, and reasoning video tasks. VURB comprises 2,100 preference pairs with long CoT traces averaging 1,143 tokens, and adopts majority voting to mitigate position bias~\cite{positionbias}. 
Evaluation on VURB reveals that existing reward models perform poorly on video understanding preference judgment: most open-source models fall below 55\% accuracy and models trained solely on text or image-text preferences fail to transfer to video understanding reward tasks. We attribute this primarily to the lack of dedicated video understanding preference data, which is essential for robust video understanding preference judgment. To this end, we construct \textbf{Video Understanding Preference Dataset} (VUP-35K) via a fully automated pipeline. Building on this data, we train \textbf{VideoDRM} and \textbf{VideoGRM} (a discriminative and a generative reward model), which achieve state-of-the-art performance on VURB and VideoRewardBench. 
Further experiments show that VUP-35K strengthens video understanding and reasoning, while both models provide substantial gains under best-of-$N$ test-time scaling.

In summary, our main contributions are as follows:
\begin{itemize}
    \item We propose \textbf{VURB}, a robust video understanding preference benchmark with long CoT traces and majority-voting evaluation across comprehensive video tasks.
    \item We construct \textbf{VUP-35K}, a large-scale video understanding preference dataset using a fully automated pipeline.
    \item We develop \textbf{VideoDRM} and \textbf{VideoGRM}, which achieve state-of-the-art on VURB and VideoRewardBench~\cite{videorewardbench} and improve best-of-$N$ test-time scaling.
\end{itemize}

\section{Related Works}
\label{sec:related}

\subsection{Multimodal Reward Modeling}
Multimodal reward models provide reward signals for reinforcement learning or data filtering~\cite{rm,xiong2025llava,xiong2026phycritic,xiong2025multi}. They can be categorized along two axes: (1) \emph{pointwise} models that score a single output independently versus \emph{pairwise} models that rank multiple response candidates, and (2) \emph{discriminative} training that attaches a scalar head optimized with the Bradley-Terry loss~\cite{rankingloss} versus \emph{generative training} that fine-tunes MLLMs to produce textual critiques or rankings via supervised fine-tuning (SFT) or reinforcement learning (RL). Under this taxonomy, SkyWork-VL-Reward~\cite{skyworkvlreward} trains a discriminative pointwise model for image-text understanding; Flex-Judge~\cite{flexjudge} and LLaVA-Critic-R1~\cite{llavacritic} perform generative training with SFT and RL respectively; UnifiedReward~\cite{unifiedreward} mixes pointwise and pairwise data to support both settings; UnifiedReward-Thinking~\cite{unifiedrewardthinking}, R1-Reward~\cite{r1reward}, and OmniReward~\cite{omnireward} further scale preference data across modalities and training stages. However, none of these works target the video understanding domain due to the lack of high-quality video understanding preference data. 

\subsection{Multimodal Reward Benchmarks}
Existing benchmarks for evaluating multimodal reward models (MRMs) primarily target text or image modalities~\cite{vlrewardbench,rewardbench,multimodalrewardbench}. Recent efforts have begun to extend evaluation to video understanding, but still suffer from critical limitations. MM-RLHF-RewardBench~\cite{mmrlhf} and JudgeAnything~\cite{judgeanything} are limited in sample coverage. OmniRewardBench~\cite{omnireward} and VideoRewardBench~\cite{videorewardbench} offer broader scope but share two key shortcomings: (1)~their preference responses lack chain-of-thought reasoning, yielding very short average lengths (133 and 104 tokens, respectively) that are misaligned with the prevailing \textit{think-then-answer} paradigm in video understanding~\cite{videocot}, which limits practical utility; and (2)~their single-pass evaluation protocol is highly susceptible to position bias~\cite{positionbias}, undermining robustness. In contrast, our VURB is a holistic video-understanding reward benchmark that spans comprehensive evaluation dimensions, curates preference data with long chain-of-thought~\cite{cot} responses (averaging 1,143 tokens), and employs a majority-voting protocol for more robust and reliable assessment.

\begin{figure*}[t]
\centering
\includegraphics[width=\textwidth]{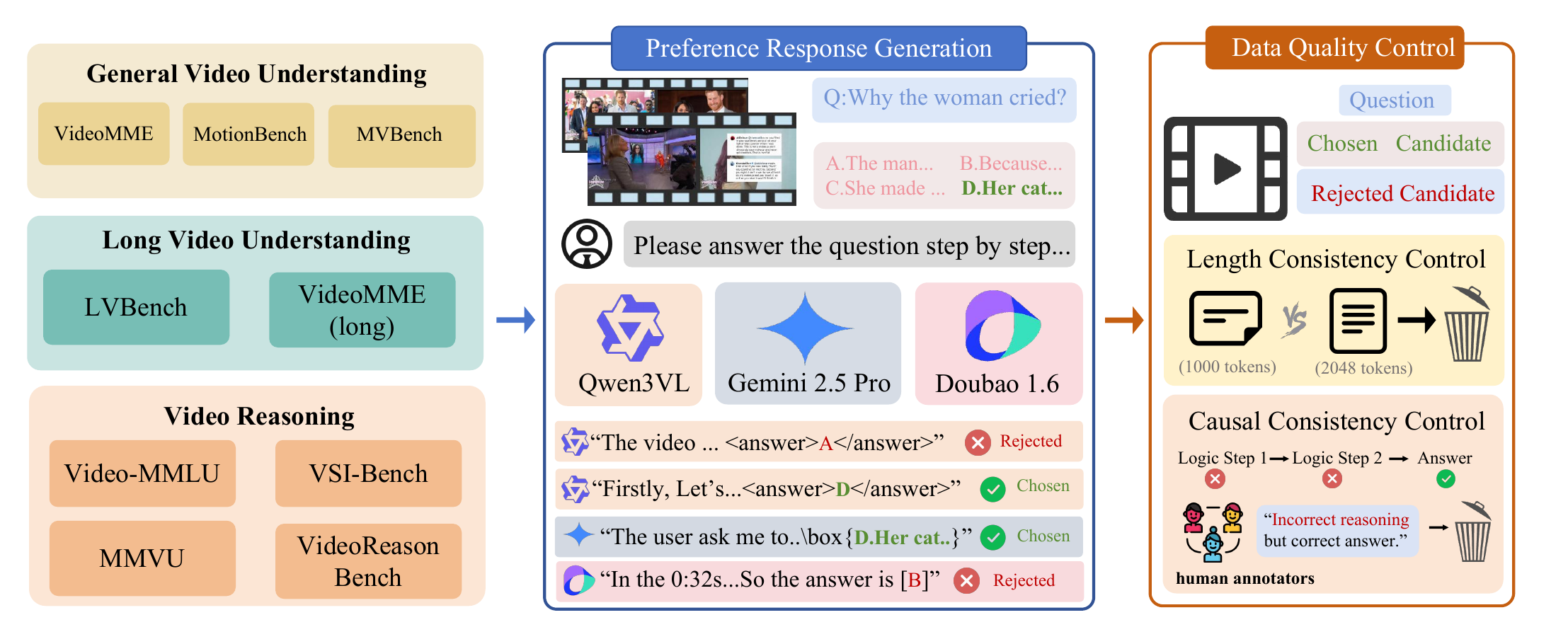}
\caption{\textbf{Overview of the VURB construction pipeline.} We first sample prompts from multiple sources to ensure diversity and broad task coverage (Sec.~\ref{PromptSampling}). We then generate preference responses using five multimodal models (Sec.~\ref{ResponseGeneration}). Finally, we apply a two-stage data quality control procedure (Sec.~\ref{datafiltering}), including (1) length consistency control, which filters response pairs with relative length differences larger than 25\%, and (2) causal consistency control, which removes samples where reasoning and final answers are inconsistent.}
\label{fig4}
\end{figure*}

\section{Video Understanding Reward Benchmark}
\label{sec:VURB}

In this section, we describe the VURB construction pipeline illustrated in Fig.~\ref{fig4}. Following prior work~\cite{vlrewardbench}, VURB is formulated as preference tuples $(x, y_c, y_r)$, where $x$ denotes a video--prompt query and $(y_c, y_r)$ denote the chosen and rejected responses, respectively. Compared with VideoRewardBench~\cite{videorewardbench}, where responses may contain only final answers, our benchmark features long chain-of-thought responses in all preferences. In the following subsections, we present prompt sampling, response generation, and quality control. Finally, we report benchmark statistics.

\subsection{Multi-source Prompt Sampling}
\label{PromptSampling}
To ensure diversity and broad coverage, we adopt a multi-source strategy and collect prompts from three domains: general video, long video, and video reasoning.

\noindent\textbf{General Video Understanding.}
This domain covers general scenarios, designed to evaluate whether multimodal reward models can assess long-CoT responses in common settings. We sample prompts from three benchmarks: VideoMME~\cite{videomme}, MotionBench~\cite{motionbench}, and MVBench~\cite{mvbench}. VideoMME~\cite{videomme} spans 6 visual domains and 30 sub-domains; we use only its short and medium subsets in this domain and reserve the long subset for Long Video Understanding. MotionBench~\cite{motionbench} evaluates fine-grained motion understanding, and MVBench~\cite{mvbench} offers broader temporal evaluation. All sampled questions are in multiple-choice question (MCQ) format.

\noindent\textbf{Long Video Understanding.}
Long video understanding is critical for modern video systems. We sample prompts from LVBench~\cite{lvbench} and the long-video split of VideoMME~\cite{videomme}, covering videos spanning tens of minutes to hours. This domain evaluates the judging capability of multimodal reward models under long-context video inputs. The sampled questions are also in MCQ format.

\noindent\textbf{Video Reasoning Task.}
This domain focuses on benchmarks requiring fine-grained logical reasoning. Responses typically involve complex CoT and multi-step inference, leading to longer outputs and posing greater challenges for multimodal reward models, which must identify flawed reasoning steps. Specifically, we include knowledge-oriented MMVU~\cite{mmvu} and Video-MMLU~\cite{videommlu}, spatial VSI-Bench~\cite{vsibench}, and vision-centric VideoReasonBench~\cite{videoreasonbench}. The sampled questions include both MCQ and short-form open-ended formats.

\subsection{Preference Response Generation}
\label{ResponseGeneration}
To ensure diverse CoT patterns, we generate responses using five models: Qwen3VL-8B (Instruct/Thinking)~\cite{Qwen3VL}, Qwen3VL-32B-Thinking~\cite{Qwen3VL}, Seed1.6-VL-Thinking~\cite{seedvl}, and Gemini-2.5-Pro~\cite{gemini}. Most are ``thinking-enabled'' variants, which produce longer, more complex traces, making the benchmark reflective of real-world usage.

As shown in Fig.~\ref{fig4}, for each prompt, we ask all models to generate CoT responses. We then extract each final answer and employ Qwen3-4B~\cite{qwen3} to verify it against the ground truth. To maintain difficulty and comparability, samples for which all responses are correct or all are incorrect are discarded. From the remaining samples, one correct response is selected as chosen and one incorrect response as rejected, forming a preference pair. The CoT prompt is provided in the Appendix.

\subsection{Data Quality Control}
\label{datafiltering}
To ensure quality, we apply a two-stage filtering strategy. In the first stage, to enable fairer comparisons and reduce reward hacking~\cite{lengthbias} (e.g., longer responses being more likely correct or preferred), we apply \textit{length consistency control} to response verbosity. Specifically, we only pair responses that satisfy
$\frac{|l_1-l_2|}{\min(l_1,l_2)} < \tau$
where $l_1$ and $l_2$ are response word counts and $\tau$ is the maximum relative length difference. We set $\tau=0.25$ to balance control and diversity.

In the second stage, we define valid CoT as reasoning that supports the final answer. Accordingly, three human annotators review all responses and apply \textit{causal consistency control} to filter out samples where reasoning and answer correctness conflict. Any sample questioned by any annotator is discarded. This reduces the preference set from 2,450 to a final 2,100 pairs.

\subsection{Benchmark Statistics}
\label{DataStatistics}
As shown in Table~\ref{tab:vurb-statics}, the benchmark comprises 2,100 preference pairs from eight video benchmarks, distributed across three dimensions for broad coverage, and encompasses 440 unique videos lasting from under 1 to 60 minutes. Responses are sampled from five models, each including explicit chain-of-thought reasoning with an average length of 1,143 tokens and a maximum of 15,378.

\suppressfloats[t]
\begin{table}[t]
  \centering
  {\fontsize{9pt}{10pt}\selectfont
\begin{tabular}{lr}
\toprule
\textbf{Statistic} & \textbf{Number} \\
\midrule
Total Preference Pairs & 2100 \\
\hspace{1em}- General Video Understanding & 797 (38.0\%)\\
\hspace{1em}- Video Reasoning & 858 (40.8\%)\\
\hspace{1em}- Long Video Understanding & 445 (21.2\%)\\
Source Datasets Pairs & \\
\hspace{1em}- VideoMME & 506 \\
\hspace{1em}- MotionBench & 135 \\
\hspace{1em}- MVBench & 256 \\
\hspace{1em}- LVBench & 345 \\
\hspace{1em}- Video-MMLU & 303 \\
\hspace{1em}- VSI-Bench & 111 \\
\hspace{1em}- MMVU & 215 \\
\hspace{1em}- VideoReasonBench & 229 \\
\midrule
Total Videos & 440 \\
\hspace{1em}- Short Videos ($\leq 1$ min) & 214 (48.6\%) \\
\hspace{1em}- Medium Videos (1 $\sim$ 5 min) & 158 (36.0\%) \\
\hspace{1em}- Long Videos ($> 5$ min) & 68 (15.4\%) \\
\midrule
Question Word Count (avg/max) & 46/298 \\
Chosen Response Word Count (avg/max) & 1143/15378 \\
Rejected Response Word Count (avg/max) & 1159/16086 \\
\bottomrule
\end{tabular}}

  \caption{Overview statistics of VURB.}
  \label{tab:vurb-statics}
\end{table}

\section{Video Understanding Reward Model}
\label{sec:VURM}

In this section, we describe the construction of training data, and the design and training of two reward models. We construct VUP-35K, a preference dataset for reward modeling. To provide complementary supervision, we incorporate 210K image-text preference pairs. Built on this corpus, we introduce two models: \textbf{VideoDRM}, a discriminative model, and \textbf{VideoGRM}, a generative model.

\subsection{Training Data Construction}
Prior work~\cite{mmrlhf,omnireward,llavacritic} has introduced various forms of human preference data, mainly covering text-only and image-text understanding. However, high-quality video preference data with long CoT reasoning remains scarce due to construction complexity and annotation cost. To address this gap, we construct \textbf{VUP-35K}, a dataset of 35K high-quality reasoning preference pairs for video understanding.

We adopt a human-annotation-free, automatic construction pipeline. For each verifiable sample (multiple-choice or open-ended questions with ground truth), we \emph{rollout} each prompt $N$ times using $Y$ candidate models, yielding $Y \times N$ reasoning trajectories. Qwen3-4B~\cite{qwen3} extracts the final answer from each trajectory and compares it against the ground truth: correct trajectories serve as chosen responses and incorrect ones as rejected. Qwen3VL-30B-A3B-Thinking~\cite{Qwen3VL} then verifies causal consistency between the reasoning and final answer in each trajectory, discarding inconsistent samples. Additionally, we remove samples with excessive length disparity or extreme difficulty (i.e., all trajectories correct or all incorrect).

In practice, we apply this pipeline to the Video-R1-CoT-165K~\cite{videor1} and Video-Holmes~\cite{videoholmes} training sets, using Qwen3VL-8B-Instruct~\cite{Qwen3VL} and Qwen3VL-8B-Thinking~\cite{Qwen3VL} as candidate models with four \emph{rollouts} per sample.

Finally, to broaden general visual reward capability, we incorporate additional image-text preference pairs from LLaVA-Critic-R1~\cite{llavacritic}, R1-Reward~\cite{r1reward}, and OmniReward~\cite{omnireward}. As summarized in Table~\ref{tab:training_data}, VUP-35K and these sources together yield 245K training samples.

\begin{table}[t]
\centering
\small
\setlength{\tabcolsep}{4pt}
\renewcommand{\arraystretch}{1.30}
\resizebox{\columnwidth}{!}{%
\begin{tabular}{l l r l}
\hline
\textbf{Source} & \textbf{Subset} & \textbf{\#Size} & \textbf{Modality} \\
\hline
Full & Full & 245K & -- \\
\hline
\multirow{4}{*}{Omni-Reward} & Omni-VLFeedback & 12K & TI2T \\
 & Omni-RLAIF-V & 16K & TI2T \\
 & OmniAlign-V-DPO & 50K & TI2T \\
 & RLAIF-V & 80K & TI2T \\
\hline
\multirow{1}{*}{LLaVA-Critic-R1} & LLaVA-Critic-GRPO-dataset & 35K & TI2T \\
\hline
\multirow{1}{*}{R1-Reward} & R1-Reward-RL & 17K & TI2T \\
\hline
\multirow{2}{*}{Ours*} & Video-Holmes Preference{*} & 4K & TV2T \\
 & Video-R1 Preference{*} & 31K & TV2T \\
\hline
\end{tabular}%
}
\caption{Training data statistics. * denotes subsets constructed in this work. ``TI2T'' and ``TV2T'' denote Text-Image-to-Text and Text-Video-to-Text respectively.}
\label{tab:training_data}
\end{table}

\subsection{Video Discriminative Reward Modeling}
To develop \textbf{VideoDRM}, we adopt a standard reward modeling framework by replacing the language model head of the model's backbone with a scalar score head. Given a video input with a question and a response pair $(x, y_{\text{chosen}}, y_{\text{rejected}})$, the model learns to assign a scalar quality score $r$ to each candidate. The training objective maximizes the margin between chosen and rejected responses via the ranking loss in Eq.~\ref{eq:rm_loss}:
\begin{equation}
\mathcal{L}_{\mathrm{RM}}=-\log\big(\sigma(r_{\text{chosen}}-r_{\text{rejected}})\big),
\label{eq:rm_loss}
\end{equation}
where $r_{\text{chosen}}$ and $r_{\text{rejected}}$ denote the scores predicted by the score head for the chosen and rejected responses respectively, and $\sigma(\cdot)$ is the sigmoid function. In practice, we initialize from Qwen3VL-8B-Instruct~\cite{Qwen3VL}, freeze the vision encoder, and fine-tune the language model together with the score head. Training proceeds for 3 epochs with a cosine learning rate scheduler at a peak learning rate of $1\times10^{-6}$, weight decay of $0.1$, and a warmup ratio of $1\times10^{-5}$. Videos are sampled at 2\,FPS with a maximum resolution of $448\times448$, allowing up to 180 frames per video and a maximum sequence length of 32K tokens.

\subsection{Video Generative Reward Modeling}
For \textbf{VideoGRM}, we leverage Group Relative Policy Optimization (GRPO)~\cite{grpo} to train a generative reward model that jointly produces a preference decision and an interpretable reasoning trace. The model follows a predefined CoT template and performs $G$ \emph{rollouts} per video-query input. Each \emph{rollout} is evaluated by a binary reward function: a score of 1.0 is assigned when the model correctly identifies the ground-truth chosen response and 0.0 otherwise. By maximizing the expected reward, the model learns to align its preference judgments with ground-truth labels. We likewise initialize from Qwen3VL-8B-Instruct with a frozen vision encoder and apply GRPO for 1 epoch, using $G{=}4$ rollouts per prompt, 4 gradient steps per generation, a sampling temperature of $1.0$, KL coefficient $\beta{=}1\times10^{-3}$, and clipping parameter $\epsilon{=}0.2$. The optimizer and video input settings are the same as those of VideoDRM.

\begin{table*}[t]
  \centering
  \scriptsize
  \setlength{\tabcolsep}{4.5pt}
  \resizebox{0.95\textwidth}{!}{%
  \begin{tabular}{lcccccc>{\columncolor{n8col}}c@{\hspace{4pt}\vrule\hspace{6pt}}c@{\hspace{4pt}}c@{\hspace{3.5pt}}c@{\hspace{3.5pt}}c@{\hspace{3.5pt}}c@{\hspace{2pt}}}
\toprule
\multirow{2}{*}{Models}
  & \shortstack{General Video\\Understanding}
  & \shortstack{Video\\Reasoning}
  & \shortstack{Long Video\\Understanding}
  & \multicolumn{4}{c}{Overall}
  & \multicolumn{5}{c}{Position Bias} \\
\cmidrule(lr){2-4} \cmidrule(lr){5-8} \cmidrule(lr){9-13}
& \multicolumn{3}{c}{N=8} & N=2 & N=4 & N=6 & \textbf{N=8} & N=2 & N=4 & N=6 & N=8 & $\Delta_{8-2}$ \\
\midrule
\multicolumn{13}{l}{\textit{\textbf{Pairwise setting}}} \\
\midrule
\multicolumn{13}{l}{\textit{Commercial Models}} \\
Seed1.6-VL-Thinking & 61.6 & 60.3 & \textbf{69.0} & 54.2 & 59.3 & 61.3 & 62.6 & 24.8 & 24.0 & 22.9 & 22.0 & $-$2.8 \\
GPT-5.2 & \underline{62.6} & \textbf{63.1} & \underline{63.0} & 52.8 & \underline{59.4} & \underline{61.8} & \underline{62.9} & 24.9 & 21.6 & 20.3 & 20.1 & $-$4.8 \\
Qwen3VL-Plus & 56.5 & 61.5 & 62.7 & 49.2 & 55.5 & 58.4 & 59.9 & 27.9 & 24.8 & 22.9 & 22.6 & $-$5.3 \\
\midrule
\multicolumn{13}{l}{\textit{Open-source Models}} \\
Qwen3VL-8B-Instruct & 51.3 & 51.0 & 57.1 & 39.0 & 46.4 & 50.2 & 52.4 & 37.4 & 34.5 & 32.6 & 31.8 & $-$5.6 \\
Qwen3VL-8B-Thinking & 50.4 & 52.8 & 57.5 & 40.2 & 47.6 & 51.0 & 53.0 & 34.5 & 30.7 & 28.7 & 27.3 & $-$7.2 \\
Qwen3VL-32B-Thinking & 54.5 & 53.5 & 56.6 & 41.0 & 48.5 & 52.4 & 54.5 & 40.7 & 36.9 & 34.8 & 34.2 & $-$6.5 \\
InternVL-3.5-8B & 50.1 & 50.7 & 55.6 & 37.6 & 45.4 & 49.5 & 51.5 & 41.0 & 37.7 & 35.4 & 34.2 & $-$6.8 \\
InternVL-3.5-38B & 52.8 & 53.0 & 56.7 & 40.9 & 49.1 & 51.8 & 53.7 & 37.3 & 33.7 & 31.9 & 30.8 & $-$6.5 \\
MiMo-VL-7B-RL-2508 & 57.1 & 52.9 & 62.0 & 45.2 & 52.0 & 54.2 & 56.4 & 52.0 & 43.4 & 41.1 & 40.1 & $-$11.9 \\
\midrule
\multicolumn{13}{l}{\textit{Specialist Models}} \\
UnifiedReward-3.0-Qwen(8B)[Pairwise] & 52.9 & 55.1 & 54.1 & 40.0 & 48.8 & 51.8 & 54.0 & 43.9 & 39.4 & 38.2 & 36.7 & $-$7.2 \\
UnifiedReward-Thinking-3.0-Qwen(8B) & 52.7 & 53.0 & 59.5 & 36.8 & 47.2 & 51.6 & 54.3 & 48.2 & 40.4 & 38.0 & 35.9 & $-$12.3 \\
R1-Reward(8B) & 48.8 & 49.5 & 51.9 & 34.3 & 42.2 & 46.7 & 49.8 & 43.3 & 35.7 & 32.8 & 30.9 & $-$12.4 \\
Flex-Judge(8B) & 40.6 & 32.6 & 40.4 & 26.2 & 32.7 & 35.9 & 37.3 & 55.2 & 49.2 & 47.0 & 45.6 & $-$9.6 \\
LLaVA-Critic-R1(7B) & 46.3 & 47.5 & 51.5 & 34.8 & 42.1 & 46.2 & 47.9 & 39.6 & 36.9 & 35.4 & 34.9 & $-$4.7 \\
\textbf{VideoGRM(ours)} & 54.7 & 62.0 & 61.8 & 52.3 & 56.2 & 58.3 & 59.3 & 23.4 & 22.6 & 22.4 & 22.2 & $-$1.2 \\
\midrule
\multicolumn{13}{l}{\textit{\textbf{Pointwise setting}}} \\
\midrule
SkyWork-VL-Reward & 58.8 & 58.4 & 59.8 & \underline{58.9} & 58.9 & 58.9 & 58.9 & 0.0 & 0.0 & 0.0 & 0.0 & 0.0 \\
UnifiedReward-3.0-Qwen(8B)[Pointwise] & 56.5 & 52.0 & 55.7 & 54.5 & 54.5 & 54.5 & 54.5 & 0.0 & 0.0 & 0.0 & 0.0 & 0.0 \\
\textbf{VideoDRM(ours)} & \textbf{65.6} & \underline{62.8} & 62.2 & \textbf{63.8} & \textbf{63.8} & \textbf{63.8} & \textbf{63.8} & 0.0 & 0.0 & 0.0 & 0.0 & 0.0 \\
\bottomrule
\end{tabular}

  }
  \caption{Evaluation results on VURB. Subcategory accuracies are reported at \textbf{N=8}; Overall and Position Bias (PB) are shown across $N{=}2,4,6,8$. The gray-shaded column highlights Overall at $N{=}8$. PB is defined as $|P(\textit{prefer first})-P(\textit{prefer second})|$; lower is better (ideal $=0$). $\Delta_{8-2}$ denotes $\mathrm{PB}_{N=8}-\mathrm{PB}_{N=2}$. Pointwise models have zero PB by design.}
  \label{tab:VURB-Result}
\end{table*}

\begin{table*}[t]
  \centering
  \scriptsize
  \setlength{\tabcolsep}{2.0pt}
  \renewcommand{\arraystretch}{0.80}
  \setlength{\aboverulesep}{0pt}
  \setlength{\belowrulesep}{0pt}
  \resizebox{0.95\textwidth}{!}{%
  \begin{tabular}{lcccccccc}
\toprule
Models & \#Param & \multicolumn{2}{c}{Perception} & Knowledge & Reasoning & Safety & Overall Acc & Macro Acc \\
\cmidrule(lr){3-4}
& & long & short & & & & & \\
\midrule
\#Samples & -- & 283 & 413 & 238 & 278 & 351 & 1563 & 1563 \\
\midrule
\multicolumn{9}{l}{\textit{\textbf{Pairwise setting}}} \\
\midrule
\multicolumn{9}{l}{\textit{Commercial Models}} \\
GPT-4o & -- & 63.3 & 50.8 & 58.8 & 57.9 & 57.3 & 57.0 & 57.6 \\
Gemini-2.5-Pro & -- & 70.7 & \underline{55.9} & \textbf{65.5} & \textbf{67.3} & 62.7 & 63.6 & 64.4 \\
Claude-3.7-Sonnet (2025-02-19) & -- & 65.0 & 48.4 & \underline{63.4} & 58.3 & \underline{82.9} & 63.2 & \underline{63.6} \\
\midrule
\multicolumn{9}{l}{\textit{Open-source Models}} \\
LLaVA-OneVision-72B & 72B & 64.7 & 40.9 & 59.7 & 53.6 & 73.5 & 57.6 & 58.5 \\
Qwen2.5-VL-72B & 72B & 68.9 & 48.4 & 56.7 & 52.5 & 44.7 & 53.3 & 54.3 \\
\midrule
\multicolumn{9}{l}{\textit{Specialist Models}} \\
LLaVA-Critic-72B (LLaVA-OV-72B) & 72B & 72.4 & 43.8 & 55.9 & 56.5 & \textbf{88.0} & 63.0 & 63.3 \\
UnifiedReward (LLaVA-OV-7B) & 7B & 67.1 & 48.2 & 50.4 & 45.3 & 71.2 & 56.6 & 56.5 \\
R1-Reward & 7B & 36.0 & 40.0 & 37.8 & 30.6 & 47.9 & 39.0 & 38.4 \\
Flex\_Judge & 7B & 35.0 & 35.1 & 37.0 & 37.1 & 30.2 & 34.6 & 34.9 \\
\textbf{VideoGRM(ours)} & 8B & \textbf{76.2} & 54.1 & 58.2 & \underline{60.3} & 73.5 & \underline{64.6} & \textbf{63.9} \\
\midrule
\multicolumn{9}{l}{\textit{\textbf{Pointwise setting}}} \\
\midrule
IXC-2.5-Reward & 7B & 73.5 & 51.3 & 56.3 & 52.2 & 38.7 & 53.4 & 54.4 \\
SkyWork-VL-Reward & 7B & 65.7 & 49.2 & 52.9 & 54.0 & 80.1 & 60.5 & 60.4 \\
\textbf{VideoDRM(ours)} & 8B & \underline{73.9} & \textbf{56.2} & 54.2 & 56.8 & 80.3 & \textbf{64.7} & 63.3 \\
\bottomrule
\end{tabular}
  }
  \caption{Evaluation results on VideoRewardBench. Our models show state-of-the-art performance; results of other models are from the VideoRewardBench paper~\cite{videorewardbench}. Bold / \underline{underlined} indicate the best / second best.}
  \label{tab:vrb-result}
\end{table*}

\section{Experiment}
\label{sec:Experiment}

In this section, we evaluate multimodal reward models on VURB across two input--output paradigms: Pairwise setting, where reward models take a chosen and a rejected response as input and output their relative ranking, and Pointwise setting, where models score each response independently. We further assess our models on VideoRewardBench~\cite{videorewardbench} to demonstrate their reward-judging capabilities.

\subsection{Evaluation Models}
For the pairwise paradigm, we evaluate generative reward models in three groups: (1) commercial models, including GPT-5.2~\cite{gpt5}, Seed1.6-VL-Thinking~\cite{seedvl} and Qwen3VL-Plus~\cite{Qwen3VL}; (2) open-source models, covering Qwen3VL~\cite{Qwen3VL} series (8B-Instruct, 8B-Thinking, and 32B-Thinking), InternVL-3.5 (8B and 38B)~\cite{internvl3d5} and MiMo-VL-7B-RL-2508~\cite{mimovl}; and (3) specialist models, including UnifiedReward-3.0-Qwen[Pairwise]~\cite{unifiedreward}, UnifiedReward-Thinking-3.0-Qwen~\cite{unifiedrewardthinking}, R1-Reward~\cite{r1reward}, Flex-Judge~\cite{flexjudge}, LLaVA-Critic-R1~\cite{llavacritic}, and our VideoGRM. For pointwise models, we evaluate UnifiedReward-3.0-Qwen[Pointwise]~\cite{unifiedreward}, SkyWork-VL-Reward~\cite{skyworkvlreward} and our VideoDRM.

\subsection{Implementation Details} 
Unlike VideoRewardBench~\cite{videorewardbench} and OmniRewardBench~\cite{omnireward}, which use a single pass and are thus more susceptible to position bias~\cite{positionbias}, we adopt Majority Voting similar to JudgeAnything~\cite{judgeanything}. For each preference sample, the model performs $N$ independent judgments, randomly swapping the order of chosen and rejected responses. A sample is correct only if the chosen response is selected more often across the $N$ trials. To quantify residual position bias, we report \textbf{Position Bias (PB)} as $\mathrm{PB}=|P(\textit{prefer first})-P(\textit{prefer second})|$, the absolute gap between favoring the first versus second slot across swapped-order trials. An unbiased model yields PB near zero; larger PB indicates stronger position susceptibility. Pointwise models score each response independently and thus have $\mathrm{PB}=0$ by design. We also report $\Delta_{8-2}=\mathrm{PB}_{N=8}-\mathrm{PB}_{N=2}$ to measure how larger $N$ mitigates this bias. We report results with $N{=}8$ to balance robustness and cost, using Overall Accuracy as the primary metric. Specialist models use prompts from their official repositories; all others use a unified template. For video inputs, we adopt model-specific frame sampling: open-source models use duration-adaptive FPS, resolution, and max-frame caps (up to 240 frames), while closed-source models follow API limits (e.g., up to 120 for Qwen3VL-Plus). Detailed prompts, frame schedules, and inference settings are in the Appendix.

\subsection{Evaluation Results on VURB}
Table~\ref{tab:VURB-Result} reports results on VURB, from which we draw the following observations.

\noindent\textbf{Existing reward models perform suboptimally on VURB.} Performance remains limited across categories: the best open-source generative model (MiMo-VL-7B-RL-2508) and pointwise model (SkyWork-VL-Reward) reach only 56.4\% and 58.9\% overall accuracy. This narrow margin over the 50\% random baseline underscores the inadequacy of current open-source models for video understanding reward tasks. Moreover, GPT-5.2 (62.9\%) still struggles, confirming that judging video understanding preferences remains a bottleneck for state-of-the-art MLLMs and motivating video-specific reward modeling.

\noindent\textbf{Inadequacy of Cross-Modality Preference Transfer.} Specialist models trained only on text or image-text data degrade substantially on video understanding reward tasks. Flex-Judge, relying on reasoning-oriented text preferences, reaches only 37.3\% accuracy. R1-Reward and LLaVA-Critic-R1, strong on image-text reward tasks~\cite{r1reward,llavacritic}, fall below 50\% on VURB, confirming that static or textual preference supervision cannot fully transfer to video understanding reward judgment.

\noindent\textbf{Superior Performance of VideoDRM and VideoGRM.} 
Via discriminative and generative reward training, VideoDRM and VideoGRM substantially improve over their Qwen3VL-8B-Instruct backbone (52.4\%). VideoDRM reaches a state-of-the-art overall accuracy of 63.8\% (+11.4\%), surpassing GPT-5.2 (62.9\%) and the largest Qwen3VL variant by 4\%. VideoGRM reaches 59.3\% (+6.9\%), becoming the leading open-source generative reward model, surpassing Qwen3VL-32B-Thinking by 4.8\% and closing the gap with Qwen3VL-Plus, showing the effectiveness of both training paradigms.

\noindent\textbf{Position Bias Mitigation by Increasing $N$.}
As shown in Table~\ref{tab:VURB-Result}, larger $N$ reduces Position Bias in the pairwise setting, while Overall accuracy rises then plateaus (notably from $N{=}2$ to $N{=}4$, with smaller gains beyond $N{=}6$). This confirms that majority voting over randomly swapped trials mitigates position bias and yields more stable evaluations.

\subsection{Evaluation Results on VideoRewardBench}
To verify generalization across video scenarios, we evaluate VideoDRM and VideoGRM on VideoRewardBench~\cite{videorewardbench}. As shown in Table~\ref{tab:vrb-result}, both achieve state-of-the-art performance: VideoDRM reaches 64.7\% overall accuracy, followed by VideoGRM at 64.6\%. Both outperform proprietary baselines such as Gemini-2.5-Pro (63.6\%) and larger open-source models like LLaVA-OneVision-72B (57.6\%) or Qwen2.5-VL-72B (53.3\%). Their consistent superiority underscores the value of specialized video reward modeling over general-purpose multimodal pre-training.

\begin{figure*}[t]
  \centering
  \includegraphics[width=0.70\textwidth]{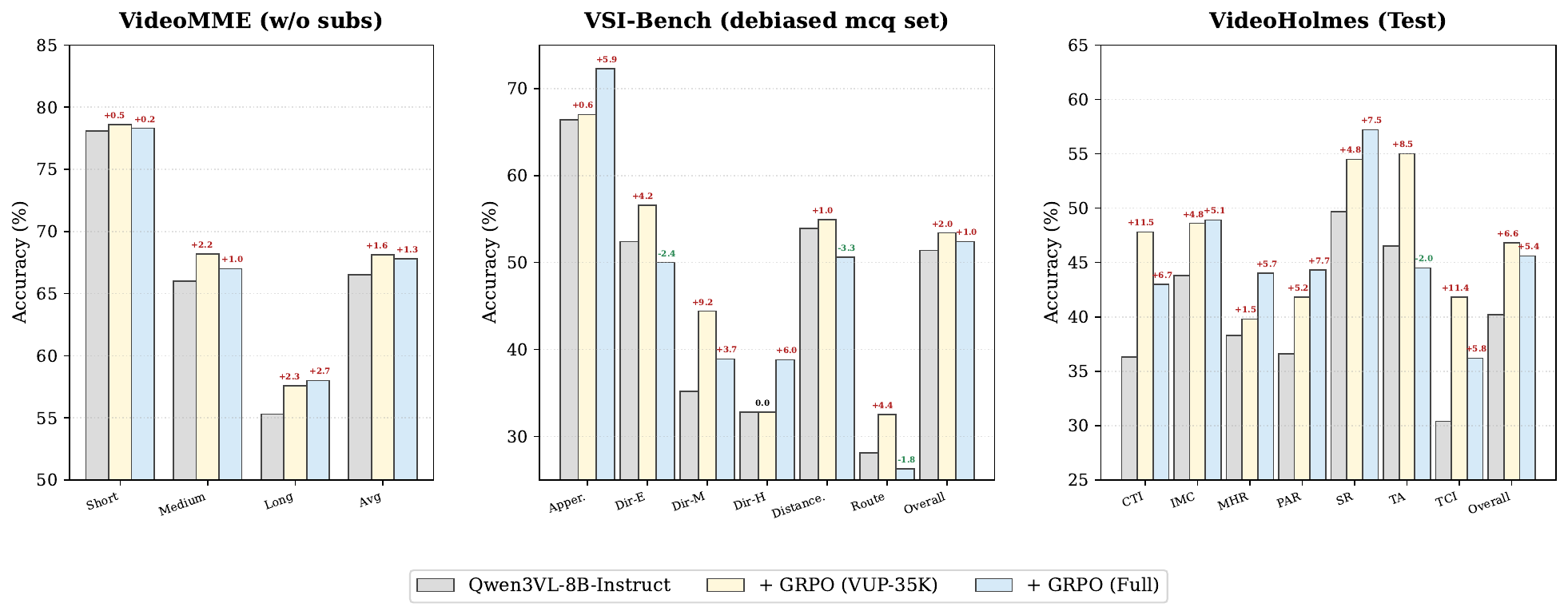}
  \caption{\textbf{Generative reward training improves downstream video understanding.} Accuracy on VideoMME (without subtitles), debiased VSI-Bench MCQ, and the Video-Holmes test set; VUP-35K training improves Video-Holmes by 6.6\%.}
  \label{fig:downstream_eval}
\end{figure*}

\begin{table}[ht]
  \centering
  \small
  \setlength{\tabcolsep}{5pt}
  \resizebox{0.95\columnwidth}{!}{%
  \begin{tabular}{lcccc}
\toprule
Settings & GVU & VR & LVU & \textbf{Overall} \\
\midrule
Qwen3VL-8B-Instruct (Base) & 51.3 & 51.0 & 57.1 & 52.4 \\
\midrule
\multicolumn{5}{l}{\textit{Discriminative Reward Training}} \\
+ VUP-35K & 67.5 & 63.4 & 51.7 & 62.5 \\
+ Full data mixture & 65.6 & 62.8 & 62.2 & \textbf{63.8} \\
\midrule
\multicolumn{5}{l}{\textit{Generative Reward Training}} \\
+ VUP-35K & 54.8 & 53.5 & 53.0 & 54.0 \\
+ Full data mixture & 54.7 & 62.0 & 61.8 & 59.3 \\
\bottomrule
\end{tabular}
  }
  \caption{Ablation results on VURB.}
  \label{tab:ablation-vurb}
\end{table}
\section{Analysis}
\label{sec:analysis}

In this section, we first evaluate the VUP-35K dataset through ablation studies on VURB and by analyzing the impact of generative reward training on downstream video understanding. We then assess the effectiveness of our proposed models under Best-of-$N$ test-time scaling.

\noindent\textbf{Ablation Results on VURB.} Results in Table~\ref{tab:ablation-vurb} show that adding VUP-35K raises backbone accuracy from 52.4\% to 62.5\%, with the largest gains in General Video Understanding (GVU) and Video Reasoning (VR), demonstrating the value of our video-specific preference data. Performance improves with the full data mixture, reaching 63.8\% for VideoDRM and 59.3\% for VideoGRM. Thus VUP-35K builds the core video understanding reward capability, while image-text preference data provides complementary supervision.

\noindent\textbf{Generative Reward Training on VUP-35K.}
We apply the GRPO recipe of VideoGRM on VUP-35K alone, without image-text preference data, and evaluate on VideoMME~\cite{videomme}, VSI-Bench~\cite{vsibench}, and Video-Holmes~\cite{videoholmes}. Figure~\ref{fig:downstream_eval} shows generative reward training improves the base model on most metrics. GRPO on VUP-35K yields a 6.6\% gain in accuracy on the Video-Holmes test set, showing high-quality reasoning preference signals. This aligns with observations~\cite{llavacritic} that generative reward training on high-quality preference data can improve visual understanding and reasoning. Training on VUP-35K alone yields higher gains than the Full mixture, suggesting video preferences align video understanding better than image-text pairs.

\noindent\textbf{Effectiveness of VideoGRM and VideoDRM.}
\setlength{\textfloatsep}{6pt plus 1pt minus 2pt}%
\begin{figure}[t]
  \centering
  \includegraphics[width=0.82\columnwidth]{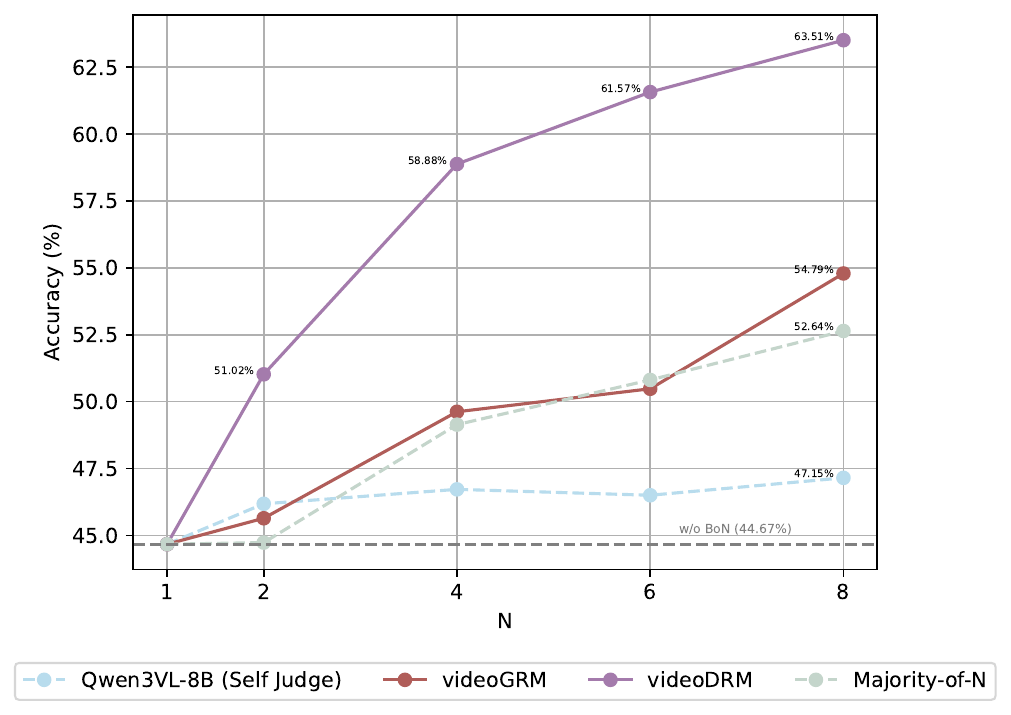}
  \caption{Best-of-$N$ results on the Video-Holmes test set (excluding all-correct or all-incorrect candidate sets).}
  \label{fig:bon-result}
\end{figure}
We conduct Best-of-$N$ experiments on the Video-Holmes test set~\cite{videoholmes} to evaluate VideoGRM and VideoDRM for inference-time selection, generating $N$ candidates per question and picking the highest-scored one. Samples with all-correct or all-incorrect candidates are excluded, as they remain unchanged under Best-of-$N$. As shown in Figure~\ref{fig:bon-result}, both outperform Self-Judge (blue) and Majority-of-$N$ (green): VideoDRM (purple) and VideoGRM (red) reach 63.51\% and 54.79\% at $N{=}8$, large gains over the 44.67\% baseline without Best-of-$N$. These results show that our reward models provide reliable preference signals for test-time reranking and improve downstream video understanding. Additional Best-of-$N$ results are provided in the Appendix.

\FloatBarrier
\section{Conclusion}
\label{sec:conclusion}

In this work, we present an end-to-end video understanding reward modeling framework featuring a more robust benchmark (VURB) and stronger reward models (VideoDRM and VideoGRM), together with the large-scale preference dataset VUP-35K. Experiments further show that existing reward models struggle on video preference judgment, while ours consistently surpass strong baselines on both VURB and VideoRewardBench, underscoring the value of specialized video reward modeling for trustworthy multimodal systems.

\bibliography{vurb}

\clearpage
\appendix

\section{Overview}
\label{sec:appendix_overview}
\begin{itemize}
    \item \textbf{Preference Response Generation} details the preference response generation procedure, including the chain-of-thought prompt template, inference parameters, the answer matching protocol, and causal consistency verification.
    \item \textbf{Training Details} expands on the training details of VideoDRM and VideoGRM, providing the full set of hyperparameters and configuration settings.
    \item \textbf{Evaluation Details} covers evaluation on Video Understanding Reward Bench (VURB), including model-specific input processing, inference parameters, prompt templates, and results under varying numbers of candidates ($N = 2, 4, 6, 8$).
    \item \textbf{Best-of-$N$ Results on VideoMME} extends the Best-of-$N$ experiments to VideoMME, further demonstrating the capability and limitations of our models as test-time reward signals.
    \item \textbf{Case Study} presents a qualitative case study on VURB.
\end{itemize}

\begin{table*}[t]
    \centering
    \scriptsize
    \setlength{\tabcolsep}{2pt}
    \renewcommand{\arraystretch}{0.78}
    \setlength{\abovecaptionskip}{4pt}
    \resizebox{0.98\textwidth}{!}{%
    \begin{tabular}{lcccccccccccc>{\columncolor{n8col}}c>{\columncolor{n8col}}c>{\columncolor{n8col}}c>{\columncolor{n8col}}c}
\toprule
\multirow{2}{*}{Models} & \multicolumn{4}{c}{General Video Understanding} & \multicolumn{4}{c}{Video Reasoning} & \multicolumn{4}{c}{Long Video Understanding} & \multicolumn{4}{c}{\cellcolor{n8col}Overall} \\
\cmidrule(lr){2-5} \cmidrule(lr){6-9} \cmidrule(lr){10-13} \cmidrule(lr){14-17}
& 2 & 4 & 6 & 8 & 2 & 4 & 6 & 8 & 2 & 4 & 6 & 8 & \cellcolor{n8col}2 & \cellcolor{n8col}4 & \cellcolor{n8col}6 & \cellcolor{n8col}8 \\
\midrule
\multicolumn{17}{l}{\textit{\textbf{Pairwise setting}}} \\
\midrule
\multicolumn{17}{l}{\textit{Commercial Models}} \\
Seed1.6-VL-Thinking & 52.8 & 57.8 & 60.3 & 61.6 & 51.9 & 57.0 & 58.8 & 60.3 & \underline{61.2} & \textbf{66.1} & \textbf{67.9} & \textbf{69.0} & 54.2 & 59.3 & 61.3 & 62.6 \\
GPT-5.2 & 51.6 & \underline{59.3} & \underline{62.0} & \underline{62.6} & 53.6 & \underline{60.2} & \underline{61.7} & \textbf{63.1} & 53.3 & 58.0 & 61.8 & \underline{63.0} & 52.8 & \underline{59.4} & \underline{61.8} & \underline{62.9} \\
Qwen3VL-Plus & 46.1 & 52.8 & 55.9 & 56.5 & 51.3 & 56.5 & 59.4 & 61.5 & 51.0 & 58.5 & 60.7 & 62.7 & 49.2 & 55.5 & 58.4 & 59.9 \\
\midrule
\multicolumn{17}{l}{\textit{Open-source Models}} \\
Qwen3VL-8B-Instruct & 38.5 & 46.5 & 49.4 & 51.3 & 37.5 & 44.9 & 48.3 & 51.0 & 42.7 & 49.2 & 55.3 & 57.1 & 39.0 & 46.4 & 50.2 & 52.4 \\
Qwen3VL-8B-Thinking & 38.8 & 47.1 & 50.1 & 50.4 & 38.6 & 45.9 & 50.3 & 52.8 & 45.8 & 51.7 & 53.9 & 57.5 & 40.2 & 47.6 & 51.0 & 53.0 \\
Qwen3VL-32B-Thinking & 41.5 & 48.8 & 52.1 & 54.5 & 37.9 & 46.0 & 50.9 & 53.5 & 46.3 & 52.6 & 56.0 & 56.6 & 41.0 & 48.5 & 52.4 & 54.5 \\
InternVL-3.5-8B & 36.5 & 44.6 & 48.5 & 50.1 & 35.4 & 43.4 & 48.4 & 50.7 & 44.0 & 50.7 & 53.6 & 55.6 & 37.6 & 45.4 & 49.5 & 51.5 \\
InternVL-3.5-38B & 40.0 & 47.2 & 49.8 & 52.8 & 38.2 & 48.0 & 51.3 & 53.0 & 47.6 & 54.4 & 56.3 & 56.7 & 40.9 & 49.1 & 51.8 & 53.7 \\
MiMo-VL-7B-RL-2508 & 47.3 & 53.2 & 55.8 & 57.1 & 39.6 & 47.8 & 50.0 & 52.9 & 52.4 & 58.2 & 59.3 & 62.0 & 45.2 & 52.0 & 54.2 & 56.4 \\
\midrule
\multicolumn{17}{l}{\textit{Specialist Models}} \\
UnifiedReward-3.0-Qwen(8B)[Pairwise] & 40.2 & 48.2 & 51.2 & 52.9 & 40.7 & 48.8 & 52.4 & 55.1 & 38.4 & 49.9 & 51.5 & 54.1 & 40.0 & 48.8 & 51.8 & 54.0 \\
UnifiedReward-Thinking-3.0-Qwen(8B) & 37.4 & 47.4 & 50.8 & 52.7 & 32.3 & 43.6 & 49.3 & 53.0 & 44.3 & 53.7 & 57.3 & 59.5 & 36.8 & 47.2 & 51.6 & 54.3 \\
R1-Reward(8B) & 34.3 & 42.5 & 44.5 & 48.8 & 33.7 & 41.3 & 47.3 & 49.5 & 35.5 & 43.6 & 49.4 & 51.9 & 34.3 & 42.2 & 46.7 & 49.8 \\
Flex-Judge(8B) & 29.7 & 35.1 & 38.5 & 40.6 & 20.9 & 28.2 & 31.2 & 32.6 & 30.3 & 37.1 & 40.2 & 40.4 & 26.2 & 32.7 & 35.9 & 37.3 \\
LLaVA-Critic-R1(7B) & 34.9 & 42.2 & 44.7 & 46.3 & 33.2 & 41.3 & 45.6 & 47.5 & 37.8 & 43.6 & 50.1 & 51.5 & 34.8 & 42.1 & 46.2 & 47.9 \\
VideoGRM(ours) & 49.0 & 51.8 & 53.8 & 54.7 & 54.1 & 58.6 & 61.4 & 62.0 & 54.6 & 59.1 & 60.2 & 61.8 & 52.3 & 56.2 & 58.3 & 59.3 \\
\midrule
\multicolumn{17}{l}{\textit{\textbf{Pointwise setting}}} \\
\midrule
SkyWork-VL-Reward & \underline{58.8} & 58.8 & 58.8 & 58.8 & \underline{58.4} & 58.4 & 58.4 & 58.4 & 59.8 & 59.8 & 59.8 & 59.8 & \underline{58.9} & 58.9 & 58.9 & 58.9 \\
UnifiedReward-3.0-Qwen(8B)[Pointwise] & 56.5 & 56.5 & 56.5 & 56.5 & 52.0 & 52.0 & 52.0 & 52.0 & 55.7 & 55.7 & 55.7 & 55.7 & 54.5 & 54.5 & 54.5 & 54.5 \\
VideoDRM(ours) & \textbf{65.6} & \textbf{65.6} & \textbf{65.6} & \textbf{65.6} & \textbf{62.8} & \textbf{62.8} & \textbf{62.8} & \underline{62.8} & \textbf{62.2} & \underline{62.2} & \underline{62.2} & 62.2 & \textbf{63.8} & \textbf{63.8} & \textbf{63.8} & \textbf{63.8} \\
\bottomrule
\end{tabular}
    }
    \caption{Evaluation results with different \textit{N} values on VURB. Bold / \underline{underlined} indicate the best / second best.}
    \label{tab:VURB-Result-N}
  \end{table*}
\section{Preference Response Generation}
\label{sec:preference_response_generation}

To construct preference pairs, we prompt all models to generate responses in a chain-of-thought (CoT) format~\cite{cot} using the template provided in Table~\ref{tab:evaluation_prompt}, with inference parameters set to a temperature of 0.95, top-$p$ of 0.95, and top-$k$ of 50. The final answer is then extracted from each response and verified against the ground truth using Qwen3-4B~\cite{qwen3} with the answer matching prompt described in Table~\ref{prompt_answer_match}. We discard any sample for which all models respond either entirely correctly or entirely incorrectly. From the remaining samples, one correct response is selected as the chosen candidate and one incorrect response as the rejected candidate, forming a preference pair. Finally, causal consistency between the reasoning and final answer in each trajectory is verified using the prompt in Table~\ref{prompt_causal_consistency}: for constructing VUP-35K (Sec.~\ref{train_details}), this check is performed automatically by Qwen3VL-30B-A3B-Thinking~\cite{Qwen3VL} to support scalable data construction, whereas for VURB, the same criterion is instead applied by three human annotators (see the Data Quality Control section of the main paper) to ensure benchmark-quality reliability. In both cases, inconsistent samples are discarded.

\begin{table*}[t]
\centering
\small
\setlength{\tabcolsep}{4pt}
\setlength{\abovecaptionskip}{4pt}
\setlength{\belowcaptionskip}{2pt}
\begin{tabular}{p{0.95\textwidth}}
\toprule
You are given the following problem: \\
<question> \\
Please think step by step and reason carefully before producing the final answer. Clearly explain your reasoning process internally, but only output the final answer. The final answer must be concise, accurate, and wrapped inside the following XML tags: \texttt{<answer> ... </answer>}. \\
\bottomrule
\end{tabular}
\caption{Prompt template for preference response generation in CoT format.}
\label{tab:evaluation_prompt}

\vspace{1.0em}
\begin{tabular}{p{0.95\textwidth}}
\toprule
You are a causal consistency evaluator for video understanding responses. Given a video, a question, a chain-of-thought reasoning trace, and a predicted final answer, determine whether the reasoning process logically supports and derives the predicted answer based on the video content and the question. \\
\\
\textbf{Evaluation criteria:} \\
1. The reasoning must be grounded in relevant visual evidence from the video (not fabricated or unrelated observations). \\
2. The reasoning steps should form a coherent logical chain that leads to the predicted answer. \\
3. The predicted answer must follow from the reasoning; judge as inconsistent if the reasoning supports a different conclusion, contains contradictions, or fails to justify the prediction. \\
\\
\textbf{Input} \\
\textbf{[Video]} <video> \\
\textbf{[Question]} <question> \\
\textbf{[Reasoning]} <reasoning> \\
\textbf{[Predicted Answer]} <prediction> \\
\\
\textbf{Output Format:} Output only a single word --- \texttt{yes} if the reasoning causally supports the predicted answer, \texttt{no} otherwise. \\
\bottomrule
\end{tabular}
\caption{Prompt template for causal consistency verification.}
\label{prompt_causal_consistency}

\vspace{1.0em}
\begin{tabular}{p{0.95\textwidth}}
\toprule
You are an answer matching evaluator. Your task is to determine whether a prediction semantically matches a ground-truth (GT) answer according to the following rules. \\
\\
1. If the GT is a choice label (e.g., A / B / C / D): judge as match if the prediction explicitly selects or refers to that label, regardless of additional explanatory text (e.g., ``C. xxx'' or ``Option C: xxx''). Case, punctuation, and extra content are ignored. Judge as not match only if it cannot be confirmed that the prediction selects the GT label. \\
2. If the GT is an open-ended answer: judge as match if the prediction expresses the same or highly similar meaning, allowing for paraphrasing, reordering, or reasonable elaboration. Judge as not match only if the prediction is semantically contradictory, clearly divergent, or missing the core information. \\
\\
\textbf{Output Format:} Output only a single word --- \texttt{yes} if matched, \texttt{no} if not. \\
\\
\textbf{GT answer:} <answer> \\
\textbf{Prediction:} <prediction> \\
\bottomrule
\end{tabular}
\caption{Prompt template for answer matching.}
\label{prompt_answer_match}

\vspace{1.0em}
\begin{tabular}{p{0.95\textwidth}}
\toprule
You are a highly skilled and impartial evaluator tasked with comparing two responses generated by a Large Multimodal Model for a given question. Start with a thorough, side-by-side comparative analysis, then choose the better response. Conclude with a single numeric choice: \\
- Output ``1'' if Response 1 is better. \\
- Output ``2'' if Response 2 is better. \\
\\
\textbf{Input} \\
\textbf{[Question]} <question> \\
\textbf{[Response 1]} <response 1> \\
\textbf{[Response 2]} <response 2> \\
\\
\textbf{Output Format:} Your detailed comparative analysis followed by the final answer in the format: \\
\begin{verbatim}
[answer]1/2[/answer]
\end{verbatim} \\
\bottomrule
\end{tabular}
\caption{Prompt template for evaluation on Video Understanding Reward Bench.}
\label{prompt_judge}
\end{table*}

\begin{figure}[t]
    \centering
    \includegraphics[width=0.85\columnwidth]{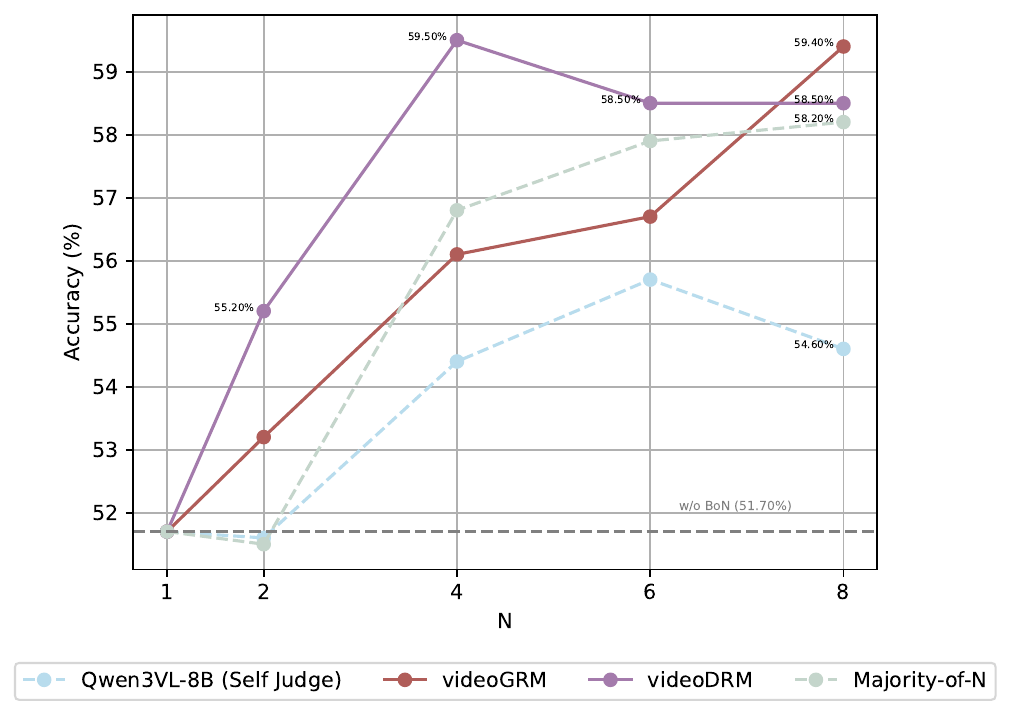}
    \caption{Best-of-$N$ results on VideoMME (excluding all-correct or all-incorrect candidate sets).}
    \label{fig:bon-result-videomme}
    \end{figure}

\section{Training Details}
\label{train_details}
Both VideoDRM and VideoGRM are trained using the Ms-Swift~\cite{msswift} framework with Qwen3VL-8B-Instruct~\cite{Qwen3VL} as the base model, keeping the vision encoder frozen throughout. For VideoDRM, we replace the LM head with a score head and fine-tune the model for 3 epochs with a cosine learning rate scheduler. For VideoGRM, we apply GRPO~\cite{grpo} training for 1 epoch, with a rollout number of 4 and steps per generation of 4, a temperature of $1.0$, $\beta$ of $1\times10^{-3}$, and $\epsilon$ of $0.2$. Both models share a learning rate of $1\times10^{-6}$, weight decay of $0.1$, and a warmup ratio of $1\times10^{-5}$. Videos are sampled at 2 FPS with a maximum resolution of $448\times448$ pixels, allowing up to 180 frames per video for both models. The maximum sequence length is set to 32K tokens for both models, and all experiments are conducted on $32\times80$GB H20 GPUs.

\section{Evaluation Details}
\label{evaluation_details}

Specialist models trained on their own preference datasets are evaluated using the prompt templates in their official repositories. All other models use a unified prompt template (Table~\ref{prompt_judge}). Each evaluation sample consists of the input question with the corresponding video frames. To ensure fair comparison, we adopt model-specific frame sampling strategies. For open-source models, frame rate (FPS), resolution, and maximum frames are adjusted based on video duration: short videos ($\le$30s) use 2.0 FPS, $512\times512$ resolution, and max frames equal to duration $\times$ FPS; medium videos (30--120s) use 2.0 FPS, $512\times512$ resolution, and up to 180 frames; long videos (120--240s) use 2.0 FPS, $448\times448$ resolution, and up to 240 frames; very long videos ($>$240s) use 1.0 FPS, $448\times448$ resolution, and up to 240 frames. InternVL models use a maximum of 12 frames at $336\times336$ resolution. Closed-source models are subject to their API input constraints: GPT-5.2~\cite{gpt5} supports up to 16 frames, Qwen3VL-Plus~\cite{Qwen3VL} supports up to 120 frames, both at $448\times448$ resolution, and Seed1.6-VL-Thinking~\cite{seedvl} follows the open-source settings above.

To mitigate position bias~\cite{positionbias}, we adopt majority voting. For each preference sample, the model performs $N$ independent judgments, randomly swapping chosen and rejected responses in each trial. A prediction is correct if the chosen response is selected more than $N/2$ times. Table~\ref{tab:VURB-Result-N} presents results for $N=2,4,6,8$. Overall, we observe the following trends: (1) In the pointwise setting, performance is unaffected by position bias, and varying $N$ has no impact on accuracy. In contrast, in the pairwise setting, model performance improves noticeably as $N$ increases---particularly from $N=2$ to $N=4$, with gains stabilizing between $N=6$ and $N=8$---indicating that position bias has a significant effect and that larger $N$ provides a more robust evaluation. (2) VideoDRM achieves the highest overall accuracy across all $N$ values, while VideoGRM remains the state-of-the-art among open-source models in the pairwise setting, demonstrating the strong performance of our models.

\section{Best-of-$N$ Results on VideoMME}
\label{sec:bon_detail}
We also conduct Best-of-$N$ experiments on VideoMME~\cite{videomme} to evaluate VideoDRM and VideoGRM in guiding inference-time selection. Samples with all-correct or all-incorrect candidates are excluded, as they remain unchanged under Best-of-$N$. As shown in Figure~\ref{fig:bon-result-videomme}, both reward models outperform the Qwen3VL-8B-Instruct~\cite{Qwen3VL} baseline across different $N$ values. Specifically, VideoDRM achieves the highest accuracy for most $N$ values, while VideoGRM shows competitive performance, particularly at $N=8$. As $N$ increases, Best-of-$N$ accuracy generally improves; however, VideoDRM experiences a slight drop from $N=4$ to $N=6$, and the Self-Judge baseline also decreases from $N=6$ to $N=8$, indicating variability across trials. At $N=8$, the gains brought by VideoGRM and VideoDRM surpass those of Self-Judge and Majority-of-$N$, demonstrating their effectiveness in inference-time selection. Nevertheless, the modest improvements compared to Majority-of-$N$ highlight a limitation of our current models, suggesting further research is needed to enhance their performance in inference-time selection.

\section{Case Study}
\label{sec:case_study}
\begin{figure*}[t]
    \centering
    \includegraphics[width=\textwidth,height=0.72\textheight,keepaspectratio]{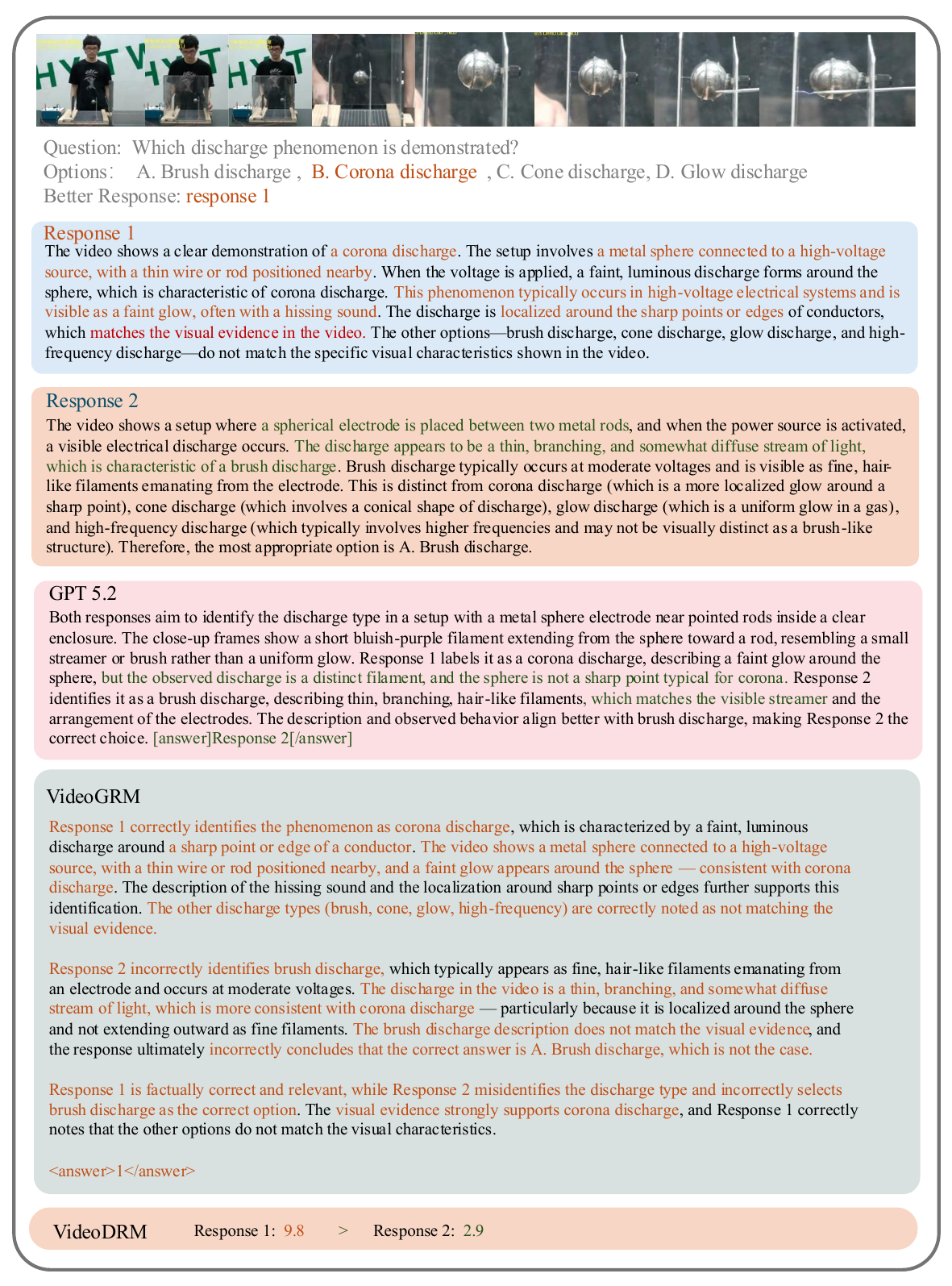}
    \caption{An example from VURB. GPT-5.2 fails to identify the salient ``sharp point'' event in the video, leading it to incorrectly select Response 2. In contrast, VideoGRM carefully analyzes the merits of Response 1 and the errors of Response 2 by grounding its judgment in visual evidence from the video, correctly identifying Response 1 as the better response. VideoDRM also correctly selects Response 1.}
    \label{fig:casestudy}
\end{figure*}
As shown in Figure~\ref{fig:casestudy}, GPT-5.2~\cite{gpt5} overlooks the salient ``sharp point'' event in the video, causing it to incorrectly favor rejected response (response 2). By contrast, VideoGRM grounds its evaluation in visual evidence, systematically identifying the strengths of Response 1 and the shortcomings of Response 2, ultimately arriving at the correct judgment. VideoDRM selects chosen response (response 1) as the better response. This case illustrates that our models are capable of making reliable and visually grounded reward judgments compared to proprietary models.

\end{document}